\documentclass[conference]{IEEEtran}
\IEEEoverridecommandlockouts
\usepackage{cite}
\usepackage{amsmath,amssymb,amsfonts}
\usepackage{algorithmic}
\usepackage{graphicx}
\usepackage{textcomp}
\usepackage{hyperref}
\usepackage{xcolor}
\def\BibTeX{{\rm B\kern-.05em{\sc i\kern-.025em b}\kern-.08em
    T\kern-.1667em\lower.7ex\hbox{E}\kern-.125emX}}
\begin{document}

\title{Exploring Kervolutional Neural Networks\\}

\author{\IEEEauthorblockN{Nicolas Perez}
\IEEEauthorblockA{\textit{Carleton University}}
}

\maketitle

\begin{abstract}
A paper published in the CVPR 2019 conference outlines a new technique called 'kervolution' used in a new type of augmented convolutional neural network (CNN) called a 'kervolutional neural network' (KNN). The paper asserts that KNNs achieve faster convergence and higher accuracies than CNNs. This "mini paper" will further examine the findings in the original paper and perform a more in depth analysis of the KNN architecture. This will be done by analyzing the impact of hyper parameters (specifically the learning rate) on KNNs versus CNNs, experimenting with other types of kervolution operations not tested in the original paper, a more rigourous statistical analysis of accuracies and convergence times and additional theoretical analysis. Code is publicly available: \href{https://github.com/NickPerezCarletonUniversity/Kervolution}{https://github.com/NickPerezCarletonUniversity/Kervolution}. 
\end{abstract}


\section{Introduction}
A kervolutional neural network (KNN) is like a traditional convolutional neural network (CNN) except it uses a 'kervolution' operation instead of a convolution operation\cite{b1}. Kervolution is like convolution, the only difference being that kernel functions are applied to filters and receptive fields before performing the dot product operation. That is where the name 'kervolutional' neural network comes from: \textbf{ker}nel con\textbf{volutional} neural network.

The convolution operation in CNNs was partly inspired by how the simple cells in the visual cortex of the human brain are used for recognizing features in images. The kervolution operation is said to be analogous not just to simple cells in the visual cortex, but also the complex and hyper complex cells that can recognize more complex patterns than the simple cells. The originally proposed paper claims that applying a kernel function on the filters and receptive fields of the convolution operation allows filters to learn nonlinear, more complex features much like how the complex and hypercomplex cells of the visual cortex do. The original paper attributes its higher reported accuracies and quicker convergence times to this. 

Based off the definition of kervolution, the convolution operation can be seen as a form of linear kervolution.

\section{Related Methods}
There are many other related methods applied to CNNs that have enabled nonlinear feature map learning or have made use of kernel functions. Out of all the described related methods in the original paper\cite{b1}, KNNs are the only method able to introduce patch-wise non-linearity while still maintaining computational complexity. These methods include:

\subsection{Quadratic Convolution}
A modified convolution operation that is quadratic with the size of the receptive field increases the capacity of filters. This method however greatly increases training complexity and training time.

\subsection{Nonlinearity through Activation Functions}
Introducing nonlinearity through activation functions (such as relu) doesn't introduce more parameters and training time, but only performs nonlinearity point wise. 

\subsection{CKN}
CKNs use kernel approximation to learn transform invariance, but can't extract non-linear features.

\subsection{SimNets}
The SimNets method adds kernel similarity layers under
convolutional layers to introduce more capacity to the model but also significantly increases the
complexity.

\subsection{Kernel Pooling}
Kernel pooling introduces more learnable non-linearity to the model but it is not able to extract patch-wise non-linear features and also increases training complexity substantially.

\section{The Kervolution Operation}
The kervolution operation makes use of the kernel trick to keep training complexity low while introducing non-linearity on the feature maps\cite{b1}. This section outlines a mathematical description of the kervolution operation which transforms CNNs into KNNs by replacing convolution with kervolution. First, a description of normal convolution is given:

$$\mathbf{conv}_{i}(x)=\left\langle x_{(i)},w\right\rangle,$$
where $x$ is the flattened input image, $i$ is the position of the receptive field (constructed using a circular shift, since $x$ is flattened) and $w$ being the filter (also a vector). Kervolution is then defined as:

$$\mathbf{kerv}_{i}(x)=\left\langle\varphi\left(x_{(i)}\right),\varphi(w)\right\rangle,$$
where $\varphi$ is a certain kernel function. The kernel trick is defined as:

$$\left\langle\varphi\left(x_{(i)}\right), \varphi(w)\right\rangle=\sum_{k} c_{k}\left(x_{(i)}^{T} w\right)^{k}=\kappa\left(x_{(i)}, w\right)$$

The kernel trick in the case of the polynomial kernel is:
$$\kappa_{\mathrm{p}}(x, w)=\sum_{j=0}^{d_{p}} c_{p}^{d_{p}-j}\left(x^{T} w\right)^{j}=\left(x^{T} w+c_{p}\right)^{d_{p}}$$

\section{Methodology}
\subsection{Methodology Used in the Original Paper}
The methodology used in the original paper relies on ablation\cite{b1}. That means that it examines certain network model configurations, and compares the performance of them with and without kervolution used while not changing any other parameters. The original paper also only uses one test set for each data set to provide accuracy scores. The two metrics used to show KNNs improved performance are highest test accuracy during training, and time to convergence where convergence in this case means achieving a target accuracy on the test data. 

The original paper shows results on a few different data sets, including MNIST and CIFAR-10 with a few different model configurations including ResNet and LeNet-5. For every experiment, the KNN gives better results on highest test accuracy during training and time to convergence.
\subsection{New Methodology Used in This Paper}
The methodology used in this paper similarly uses an ablation study, however, for both the CNN and KNN models a learning rate search is performed to see which learning rate is optimal for each model. The learning rate search is done by starting at a learning rate of $0.2$ and halving it until it is less than or equal to $0.0002$. For each learning rate the model is trained for each of $5$ k-folds. This means the new methodology uses more than one test set to provide a better analysis of metrics via a mean and confidence interval. 

Additionally, the recorded highest test accuracy of a model was only updated every time a new highest training accuracy was seen. This more simulates how a model would be developed in a real-world setting because the best model can only be determined by how well it performs on the data available to train and validate on before being sent off to do the 'real work' and final evaluation on the test data. The original paper selects the best performing model on the test data for every experiment, which is not as reflective of how well the neural network really performs. It isn't a significant change in experimentation, but something interesting to have noted. In figures~\ref{fig7} and~\ref{fig8} the correlation between training and testing accuracy can be seen.

This paper shows results on the MNIST and Fashion MNIST data sets. Fashion MNIST is considered to be a harder version of MNIST, but contains images of clothing instead of digits. The same LeNet-5 network configuration used in the original paper is used, with all of the same parameters, including milestones for learning rate decays and epochs to train. The kernels used in this paper are the polynomial kernel, which performed the best out of all the kernels tested in the original paper, and a newly proposed 'mixture of kernels' which is described in section~\ref{mixture:section}. 

\section{Mixture of Kernels}
\label{mixture:section}
A related paper showed that linearly combining the RBF and polynomial kernel functions used for a SVM can show improved performance compared to using just one of the RBF or polynomial kernels\cite{b2}. The following equation describes it:
$$\kappa_{\mathrm{m}}(x, w)=\lambda\kappa_{\mathrm{g}}(x, w) + (1-\lambda)\kappa_{\mathrm{p}}(x, w),$$
where the tunable parameter $\lambda$ is bounded between $0$ and $1$. For use in a KNN, this paper proposes that the parameter $\lambda$ be a learnable parameter wrapped in a sigmoid function to keep it bounded between $0$ and $1$:

$$\kappa_{\mathrm{m}}(x, w)=\sigma(\lambda)\kappa_{\mathrm{g}}(x, w) + (1-\sigma(\lambda))\kappa_{\mathrm{p}}(x, w)$$

This 'mixture kernel' is used the same way as the polynomial kernel in KNNs, by applying it via (two) kernel trick(s) on the receptive field and filter. The only difference is the summation component after computing the two kernel tricks.

\section{Results}
\label{results:section}
This section outlines the results given from this paper's new methodology. For the metrics in each diagram the faded colours show the $95\%$ confidence interval and the solid darker lines show the mean value. The legend of each graph shows either the maximum mean highest test accuracy and confidence interval for the highest accuracy graphs, or minimum mean convergence time and confidence interval for the convergence time graphs.
\subsection{Highest Achieved Accuracy}
\label{accuracy:subsection}
\begin{figure}[htbp]
\centerline{\includegraphics[scale=0.6]{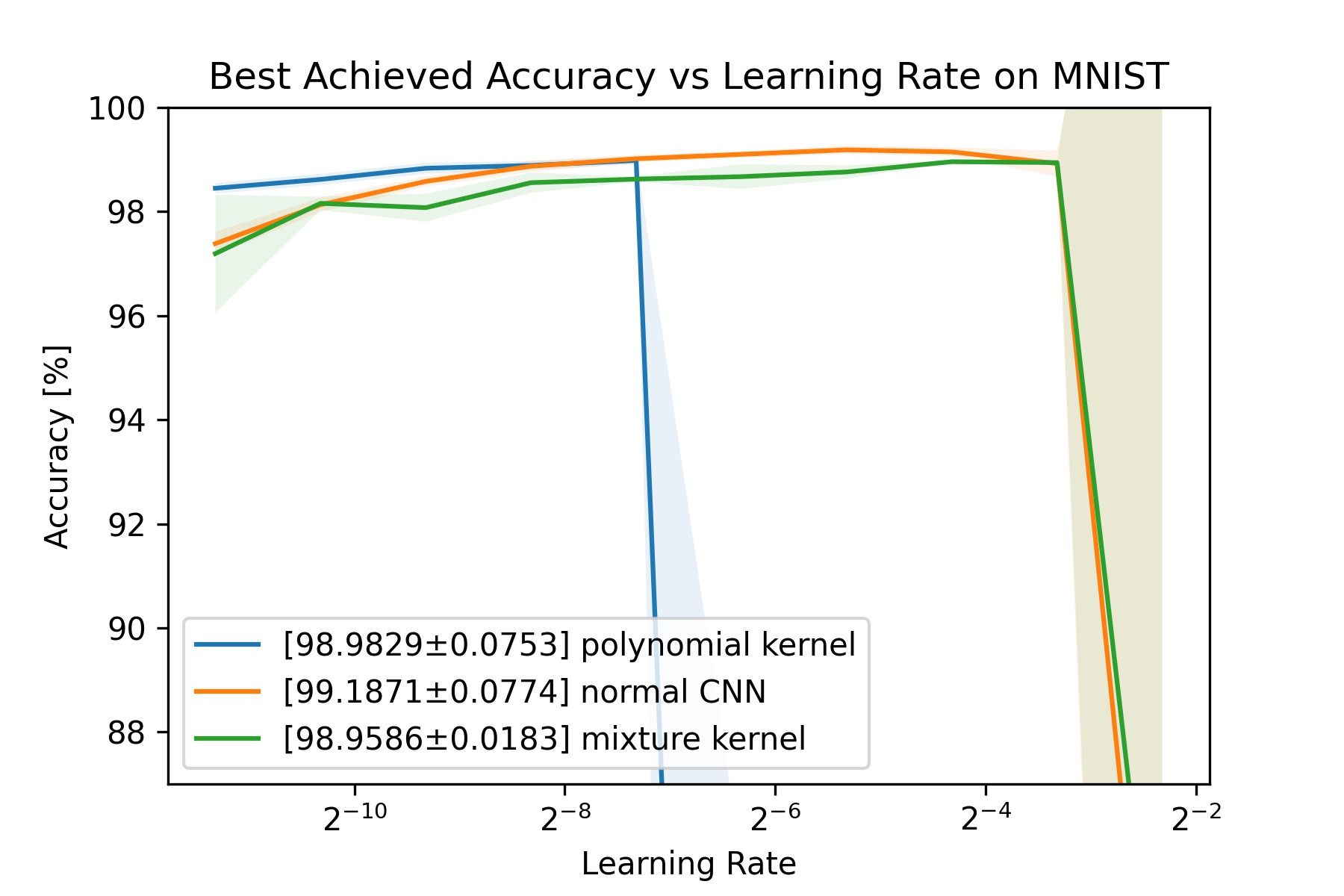}}
\caption{}
\label{fig1}
\end{figure}

\begin{figure}[htbp]
\centerline{\includegraphics[scale=0.6]{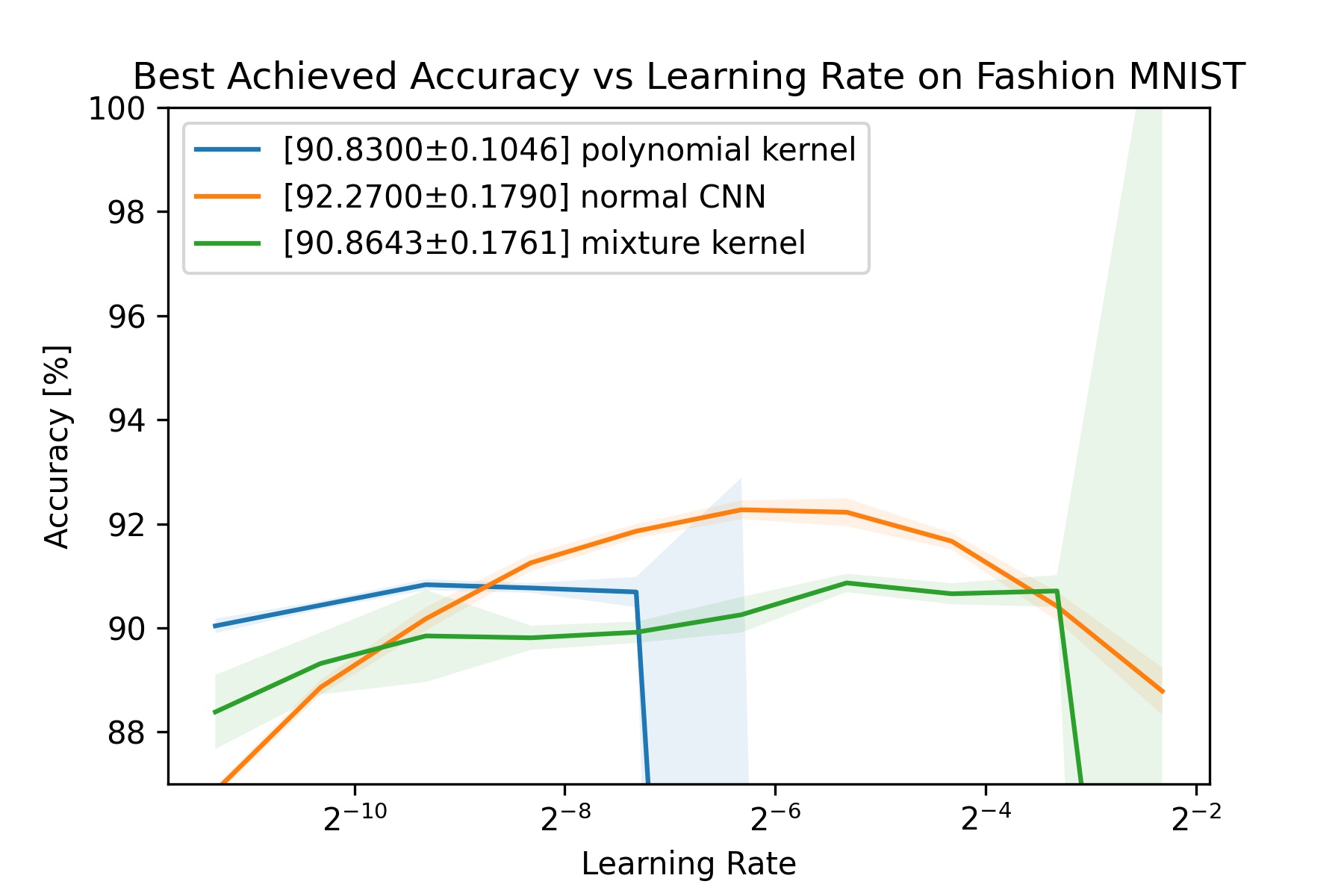}}
\caption{}
\label{fig2}
\end{figure}

See figures~\ref{fig1} and~\ref{fig2} for results showing the highest achieved accuracy vs learning rate on both the MNIST and Fashion MNIST data sets respectively. It is clearly shown that the normal CNN achieves the highest mean accuracy on both data sets. In the original KNN paper the opposite result is shown\cite{b1}, where the normal CNN achieves the lowest accuracy out of all methods on the MNIST data set. There are very wide confidence intervals attributed to outlier experiements where the network sometimes converged to $10\%$ accuracy immediately and became stuck. For pragmatic reasons, the confidence intervals given by these outlier experiments won't be taken into consideration. Taking bounds on the confidence intervals into consideration (except for the outlier experiments), the normal CNN achieves at least 95\% reproducible highest accuracy. The polynomial kernel and mixture of kernels show comparable highest achieved accuracy for both data sets.
\subsection{Time to Target Accuracy}
\begin{figure}[htbp]
\centerline{\includegraphics[scale=0.6]{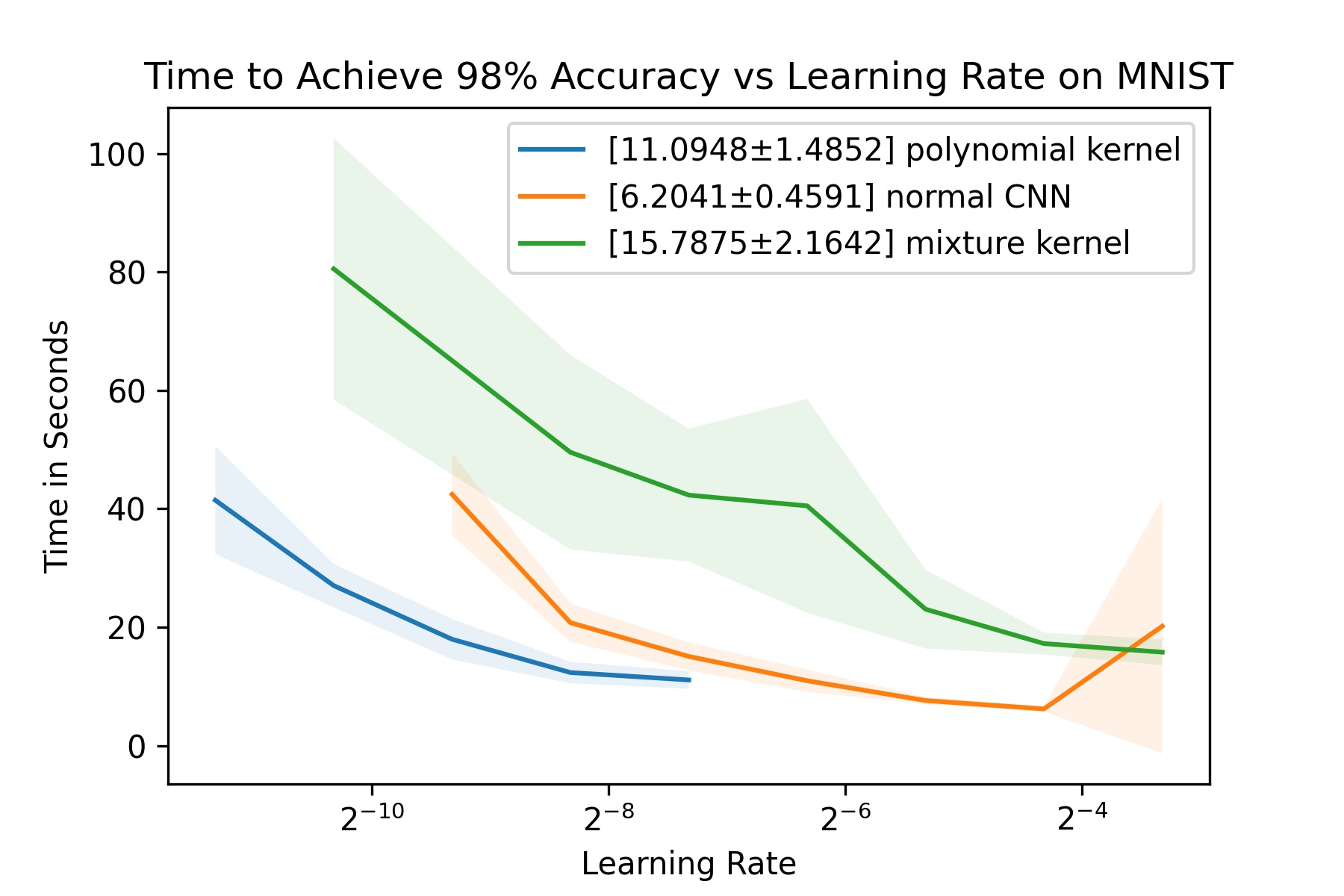}}
\caption{}
\label{fig3}
\end{figure}

\begin{figure}[htbp]
\centerline{\includegraphics[scale=0.6]{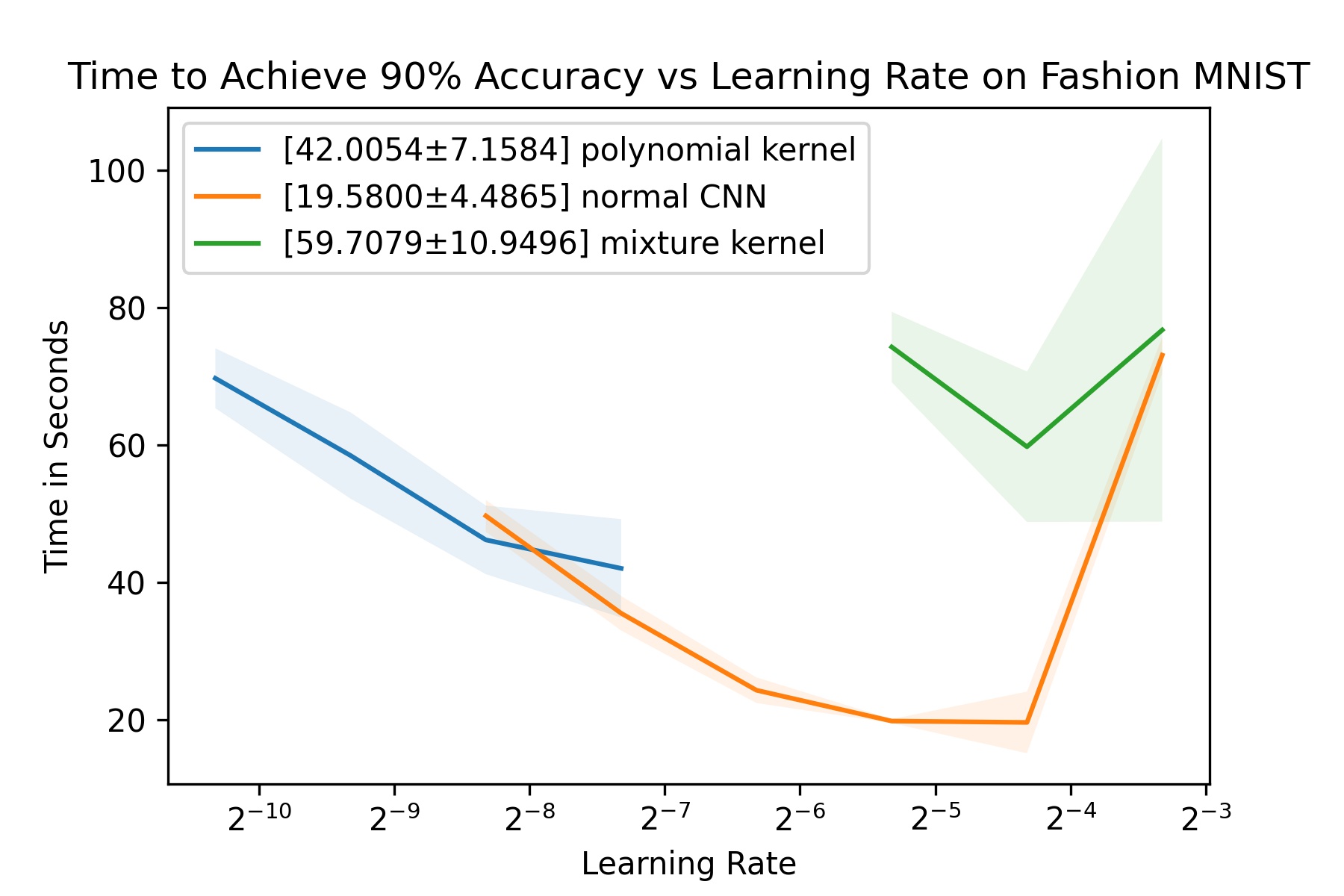}}
\caption{}
\label{fig4}
\end{figure}

\begin{figure}[htbp]
\centerline{\includegraphics[scale=0.6]{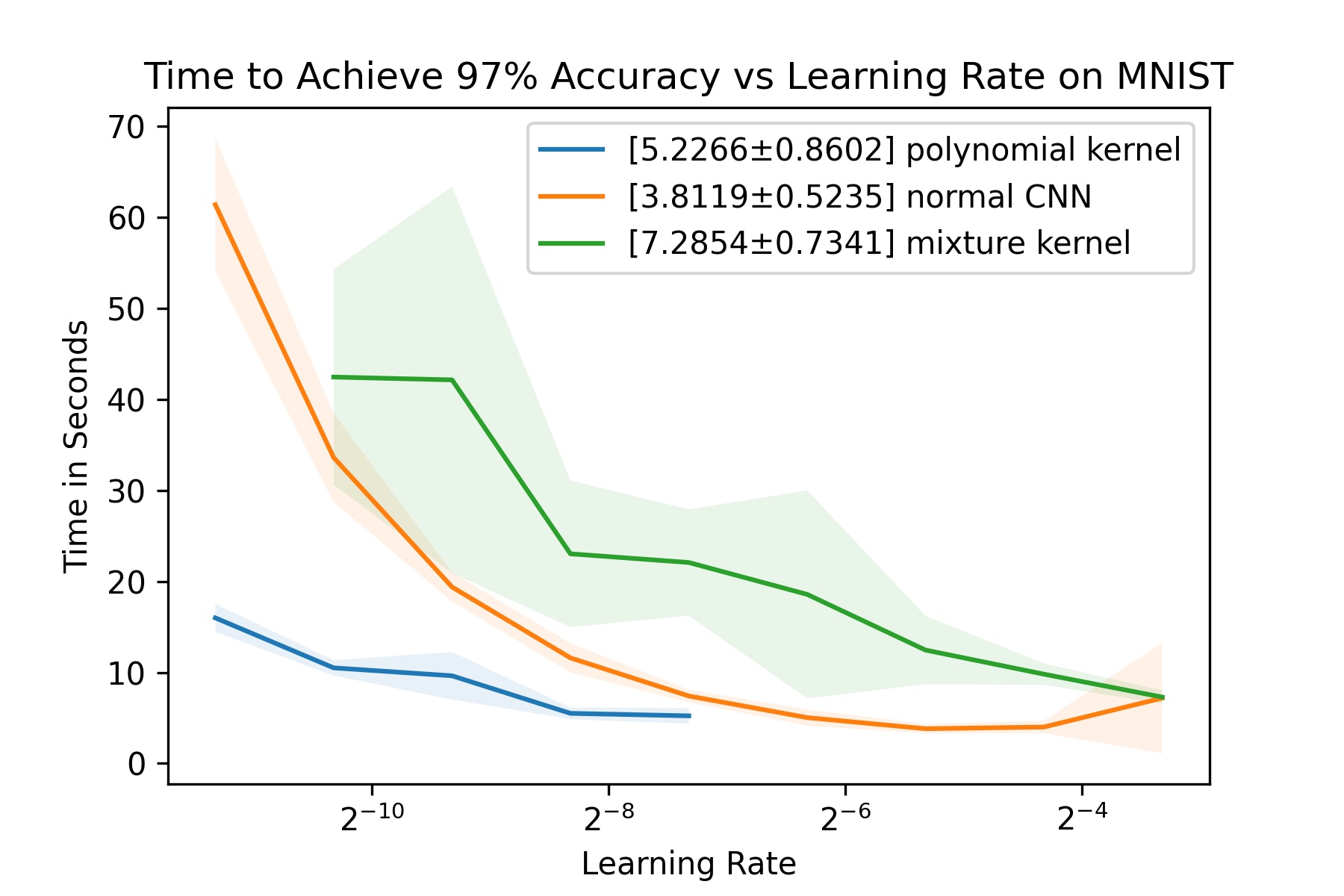}}
\caption{}
\label{fig5}
\end{figure}

\begin{figure}[htbp]
\centerline{\includegraphics[scale=0.6]{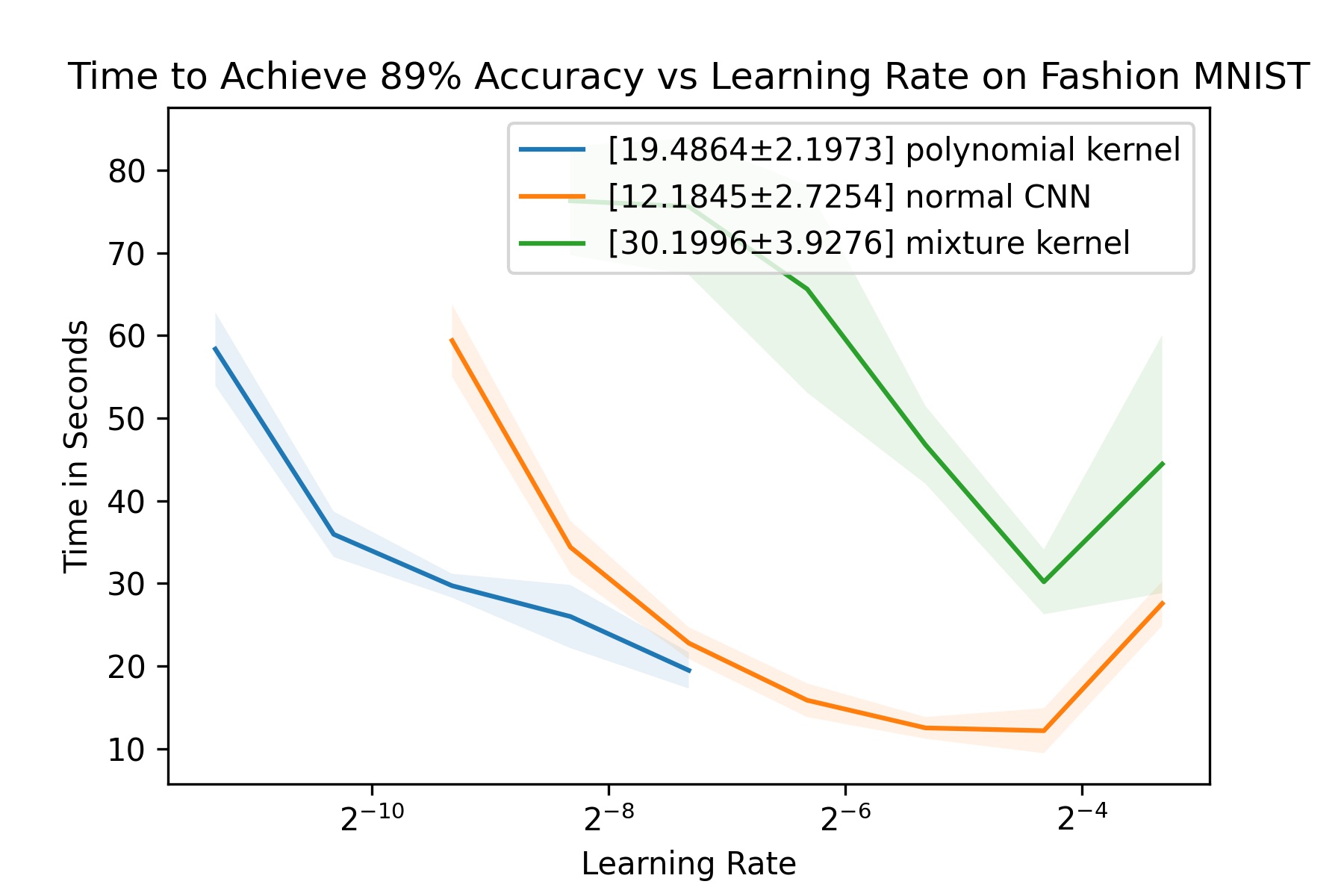}}
\caption{}
\label{fig6}
\end{figure}

See figures~\ref{fig3},~\ref{fig4},~\ref{fig5} and~\ref{fig6} describing the results for time to train to target accuracy vs learning rate on the MNIST and Fashion MNIST data sets. For each model, results for a learning rate that contain at least one fold where target accuracy was not achieved are omitted from the graph, thus alleviating the 'outlier experiment' issue described in subsection~\ref{accuracy:subsection}. The normal convolutional neural network achieved faster time to target accuracies for both data sets. The polynomial kernel and mixture kernel showed very similar results for time to target accuracy but the polynomial kernel was a bit faster. This is likely almost entirely because of the added computations from computing 2 kernels instead of one. 

\section{Analysis and Discussion}
This section contains further analysis on the results shown from section~\ref{results:section} along with additional theoretical analysis of KNNs absent from the original KNN paper.
\subsection{Learning Rates}
In the original paper the LeNet-5 model uses an initial learning rate of $0.003$ for both the KNN and CNN experiment on MNIST data. The results in section~\ref{results:section} clearly show that the learning rate used in the original paper benefitted the KNN more than the CNN. The polynomial KNN's best learning rate was $0.00625$ and the normal CNN's best learning rate was $0.025$. The polynomial kernel provided extra sensitivity to learning rates because of the higher order terms present in the filters.

The experimentation in this paper shows that it is important to be aware of the influence of hyper parameters when performing an ablation study. Given computational resources and time constraints only one hyperparameter was searched through. The learning rate was chosen as the parameter to tweak since it is perhaps the most important hyperparameter\cite{b3}. For future experiments, a much more efficient (but less exhaustive) learning rate search could be used,\cite{b4}, so that other important hyper parameter(s) can be searched through to potentially provide a more thorough analysis.
\subsection{Selecting a Kernel Operation}
One downside of the new KNN method is that there are more ways to tune a network. There are many different types of kernel functions to choose from, all that may perform better on different data sets and with different hyperparameters.

The mixture of kernels method proposed in this paper could make choosing a kernel much simpler. This could be done either through showing superior performance, or being used to evaluate a priori which kernel is most effective to use by itself.
\subsection{Kernel Operation vs Preprocessing Data}
It is interesting and potentially important to note that applying the kervolution operation is quite similar to preprocessing data. A KNN with a single kervolution operation at the input level, followed by all other kernels being strictly linear (convolution) could be equivalently constructed using preprocessing on the input data. The data would be preprocessed using the kernel function and then feeding that higher dimensional data into a CNN with all the same parameters except with a linear kernel at the input layer with an increased stride and receptive field size. The following equation illustrates this idea, where $x'$ is the transformed input data, $i'$ the new coordinate corresponding to the original receptive field's coordinate, and $w'$ the enlarged filter:

$$\mathbf{kerv}_{i}(x)=\left\langle\varphi\left(x_{(i)}\right),\varphi(w)\right\rangle=\left\langle x'_{(i')},w')\right\rangle$$

The original paper ran an ablation study where one network had a kervolutional layer at the input and then convolution for the second layer containing filters, and the other network vice versa\cite{b1}. The network that had kervolution first performed better than the network that had convolution first. This further supports the similarity between kervolution and preprocessing data. Intuitively: An effective transformation on the data is more beneficial towards the input of the network, so more layers of the network can make use of it.

\subsection{Kernel Operation Causing Overfitting}
\begin{figure}[htbp]
\centerline{\includegraphics[scale=0.6]{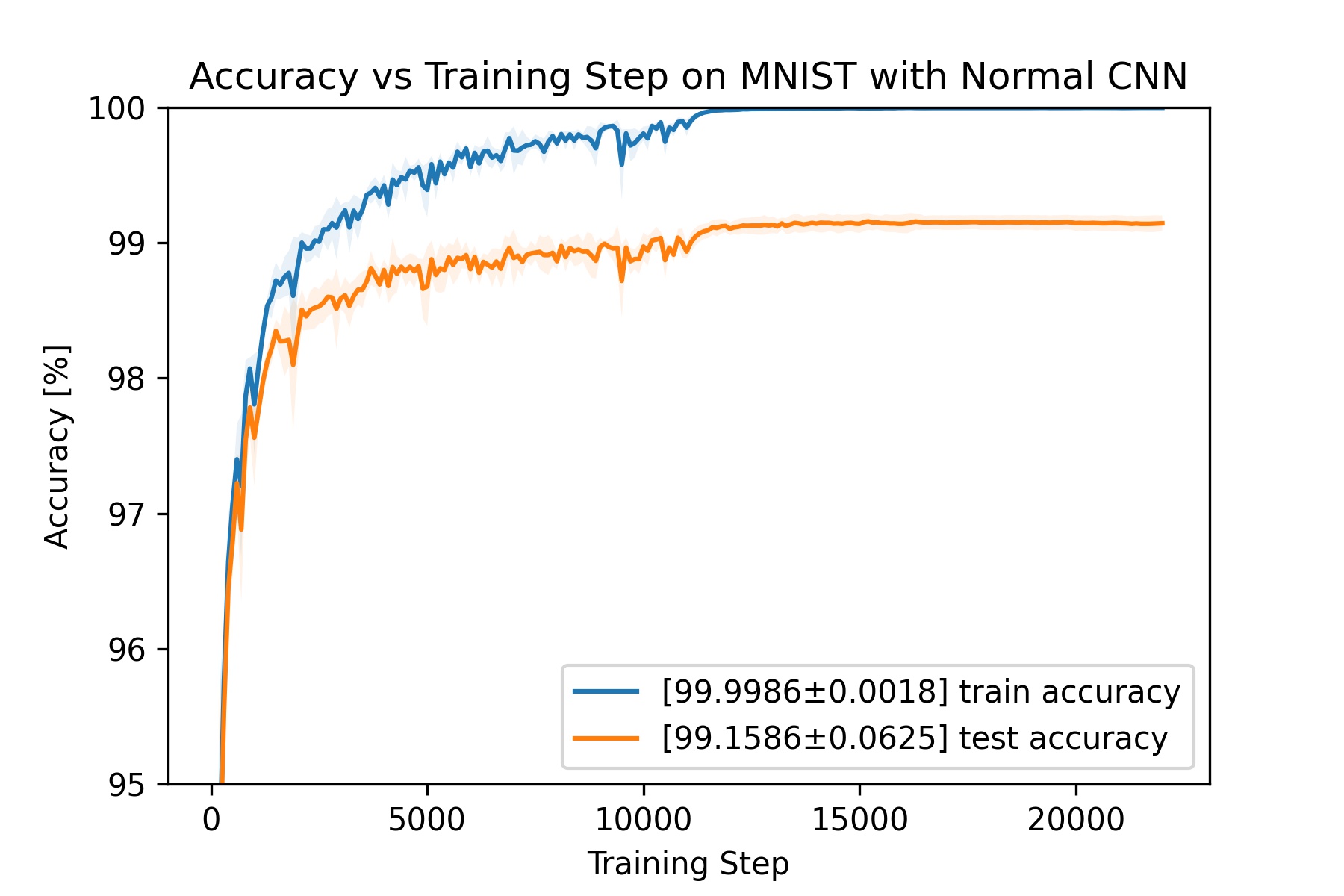}}
\caption{}
\label{fig7}
\end{figure}

\begin{figure}[htbp]
\centerline{\includegraphics[scale=0.6]{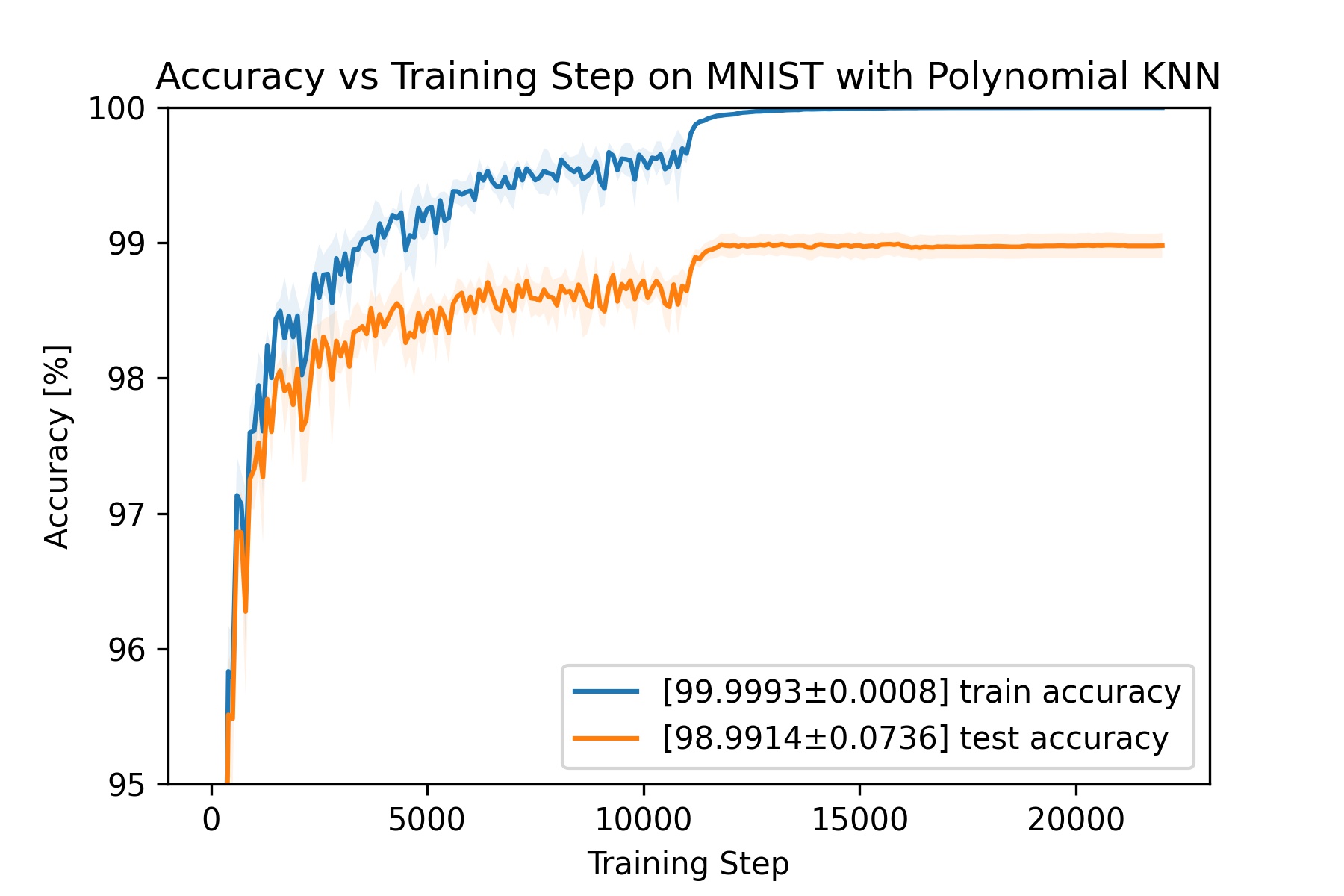}}
\caption{}
\label{fig8}
\end{figure}

In the original paper it was noted that overfitting was more present when using KNN compared to CNN\cite{b1}. The same results were found for the experimentation in this paper. So much so, that the KNN was able to achieve higher mean training accuracy than the CNN, while the CNN was achieving higher test accuracy than the KNN. The comparison of results of each model on MNIST data can be seen, where each model uses the learning rate that gave highest mean accuracy during the learning rate search. The original paper claims the overfitting showed that the LeNet-5 KNN had too high of a capacity for the MNIST data set based off of an analysis of the recorded loss. The experimentation done for this paper more effectively showed that. See figures~\ref{fig7} and~\ref{fig8}.


When kernel functions are used in SVM, a similar phenomenon happens: Increased generalization error\cite{b5}. Given enough samples, the generalization error can be minimized. Given a larger data set, the overfitting of the KNN training on MNIST could be eliminated and lead to higher test accuracies compared to CNN.
\subsection{Mixture of Kernels}
Across all experiments for each data set ($5$ folds $\times$ $10$ learning rates $= 50$ experiments), the mixture of kernels KNN achieved a higher test accuracy than the polynomial kernel KNN. For Fashion MNIST the polynomial kernel KNN achieved $91.0857\%$ accuracy while the mixture of kernels KNN achieved $91.1643\%$ accuracy. For MNIST the polynomial kernel KNN achieved $99.0643\%$ accuracy while the mixture of kernels KNN achieved $99.1429\%$ accuracy.

Although it is not conclusive that the mixture of kernels method proposed in this paper is strictly better at achieving higher accuracies than the polynomial kernel method, the results seen are encouraging for future experimentation. It is likely that putting the $\lambda$ parameter inside of the sigmoid function caused unwanted behaviour. The mixture of kernels KNN showed slightly more variant/unreliable behaviour than the polynomial KNN, such as converging to an accuracy of $10\%$. If the $\lambda$ parameter for example became close to $0$ or $1$ for a given filter, the famous 'vanishing gradient' problem may become an issue and cause this unwanted behaviour. Alternatives to this approach could include allowing the $\lambda$ parameter to change values in a more free and linear manner while having some sort of regularization on it. Due to time constraints and computational resources, this experimentation did not take place.

\subsection{Analogous Methods}
In the original paper, the use of the word 'non-linear' was mentioned many times to describe the benefit of KNN over CNN\cite{b1}. The analysis of KNN's effectiveness didn't extend much beyond 'non-linear thus improved model capacity'. It is important to note that non-linearity does not strictly mean better performance. Some analogous examples include:
\subsubsection{L1 vs L2 norm}
When using an L1 norm or an L2 norm for something such as regularizing a network, one is not strictly better than the other in all circumstances. The L2 norm's non-linearity doesn't make it superior.
\subsubsection{Applying an Activation Function to a Regression Output}
In the case of neural networks with a single regression output neuron, it isn't always advantageous to wrap the output neuron in an activation function. Introducing unnecessary non-linearity can be detrimental; introducing more non-linearity for the network to learn than necessary.

\subsection{Why Does Kervolution Work?}
Other than just providing more model capacity because of introducing non-linearity on the filters, there is another intuitive way of understanding why a KNN would be more effective than just normal CNN. Kernel functions used in SVM project data to a higher dimensional space, making it easier to separate using a hyperplane. The hyperplane in the projected space is a non-linear hyperplane in the original space. Applying a kernel function to the receptive field of a CNN might do something similar, by making features more distinct and separable. For example, in the higher dimensional space, features such as noses, eyes and ears for the purpose of face detection may be all easier to distinguish from each other, allowing filters to identify these features more accurately.

\section{Conclusion}
To conclude, this "mini paper" showed that KNNs do not strictly perform better than CNNs. More experimentation is needed to clearly justify their effectiveness, most importantly experimenting with them on larger data sets to alleviate overfitting. It was shown that that introducing the kernel on filters of a CNN is both similar to preprocessing techniques and using kernel functions in SVM. A new mixture of kernels method was proposed that showed comparable performance to the methods shown in the original KNN paper, and is potentially a starting point for further advancements.

\section{Acknowledgements}
I'd like to thank Dr. Jochen Lang from the University of Ottawa for his valuable teachings.

\end{document}